\documentclass{article}

\usepackage{newtxtext, newtxmath}
\usepackage{booktabs}

\usepackage{geometry}
\usepackage{array}
\begin{document}

\title{Hyperparameter Learning for Latent Factorization of Tensors for Representation Learning to Large-scale Dynamic Weighted Directed Network}
\author{%
\centering
\small 
\begin{minipage}{0.32\textwidth}\centering
$1^\mathrm{st}$ Yaqian Zhan\\
College of Computer and\\
Information Science\\
Southwest University\\
Chongqing, China\\
zyq000451@email.swu.edu.cn
\end{minipage}
\hfill
\begin{minipage}{0.32\textwidth}\centering
$2^\mathrm{nd}$ Jialan He\\
College of Computer and\\
Information Science\\
Southwest University\\
Chongqing, China\\
nihhu2020@email.swu.edu.cn
\end{minipage}
\hfill
\begin{minipage}{0.32\textwidth}\centering
$3^\mathrm{rd}$ Tianzhu Chen\\
College of Computer and\\
Information Science\\
Southwest University\\
Chongqing, China\\
chen56@email.swu.edu.cn
\end{minipage}
}
\date{}
\maketitle

\begin{abstract}
Large-scale dynamic weighted directed networks (DWDNs) are widely used to model time-varying interactions among nodes. Latent factorization of tensors (LFT) extracts target knowledge from DWDNs via low-rank embedding. However, similar to many machine learning models, the performance of LFT heavily depends on the selection of hyperparameters. In practice, these parameters are often tuned manually or through grid search, which requires significant computational resources and human effort. Motivated by this challenge, this paper proposes an automated hyperparameter optimization framework based on Differential Evolution (DE) for LFT (DE-LFT). The proposed method integrates DE into the training process of the LFT model to automatically learn optimal regularization parameters $\lambda_1$, $\lambda_2$ and $\lambda_3$. As a result, the model can adaptively search the hyperparameter space and improve prediction accuracy. Experimental results on four real-world datasets demonstrate that the proposed approach achieves lower MAE and RMSE compared with manually tuned baselines while reducing the need for extensive parameter tuning.
\end{abstract}

\section{Introduction}
\par In the big data era, large-scale dynamic weighted directed networks (DWDNs) have been widely used in various practical applications, such as personalized recommendation systems \cite{wu2022l1, wu2026non, wu2026federated, yu2026federated, wu2025outlier, xu2025recursion, yang2025concept, jankar2026federated}, intelligent transportation systems \cite{wang2021leveraging, wu2026multimetric, zhang2025differential, yang2025hybrid, you2024online, wu2026schemarag}, and electrical power grid infrastructures \cite{xiang2020controllability, yang2024end, yang2024adaptive,luo2025novel, wu2026multimetric}. The latent factorization of tensors (LFT) has shown strong capability in learning latent representations from DWDNs \cite{luo2021adjusting, qin2026robust, he2026modularized, he2026survey, wang2026advanced, han2026tracehg, hu2025advancing}. It models the target as a high-dimensional and incomplete (HDI) tensor and performs low-rank approximation \cite{luo2020temporal, zeng2025novel, li2026novel, valiki2026secure, wang2026graph, he2026tensor, wang2024mini, lan2026cm}.

However, similar to many machine learning models, the performance of LFT heavily depends on the selection of hyperparameters, particularly the regularization coefficients $\lambda_1$, $\lambda_2$ and $\lambda_3$ of latent feature matrices \cite{liu2025adaptive, xu2025sampling, yu2025multi, gao2025federated, ma2025review}. In practice, these parameters are often determined manually or via grid search \cite{bergstra2012random, hutter2019automated, snoek2012practical}, which is computationally expensive and inefficient. To address this issue, this paper proposes a Differential Evolution (DE) \cite{storn1997differential, hu2020grey, zhang2023novel, tong2023hybridizing} based hyperparameter learning framework for LFT. The proposed method integrates DE into the training process to automatically learn regularization coefficients and improve the model predictions accuracy.

\section{Main Work}
\par LFT models DWDNs as a third-order tensor to capture evolving node interactions. It learns three low-rank latent factor matrices $U,S,T$ that approximate observed entries via:
\[
\hat{x}_{u,s,t} = \sum_{k=1}^{K} U_{u,k} S_{s,k} T_{t,k},
\]
where $K$ is the dimensionality of the latent representation space. To optimize performance, we apply Differential Evolution (DE) for automatic hyperparameter tuning. DE is a population-based heuristic that iteratively evolves candidate solutions via mutation, crossover, and selection to find optimal regularization parameters ($\lambda_1$, $\lambda_2$ and $\lambda_3$) minimizing prediction error (MAE/RMSE). This approach adaptively explores the hyperparameter space, enhancing accuracy while reducing computational costs.

\section{Methods}
\par To evaluate the effectiveness of DE-LFT, experiments are conducted on four benchmark DWDNs. The datasets were divided into training set and test set in a 80\%: 20\% ratio. The LFT model is trained using stochastic gradient descent (SGD), while the regularization parameters are optimized using the DE algorithm. Root mean square error (RMSE) and mean absolute error (MAE) \cite{liao2025novel, han2025sgd, chen2025adaptive, yang2025fmvpci, wu2025learning} serve as evaluation metrics:
\[
\mathrm{RMSE} = \sqrt{\frac{1}{|\psi|}\sum_{x_{ijk}\in\psi} (x_{ijk}-\hat{x}_{ijk})^2},
\quad
\mathrm{MAE} = \frac{1}{|\psi|}\sum_{x_{ijk}\in\psi} |x_{ijk}-\hat{x}_{ijk}|.
\]

\section{Results}
\par The experimental results are shown in Table 1. Compared with manually tuned hyperparameters, Grid Search slightly improves the prediction performance by exploring a larger hyperparameter space. Furthermore, the proposed DE-based hyperparameter learning method consistently achieves the best performance across all datasets. This indicates that Differential Evolution can more effectively search the hyperparameter space and identify parameter combinations that yield lower prediction errors. Overall, the proposed method achieves the lowest MAE and RMSE on all four datasets.

\begin{table}[h]\centering
\caption{The Comparison Result of Rating Prediction Accuracy}
\begin{tabular}{lcccccc}
\toprule
Model & LFT & LFT & Grid Search & Grid Search & DE-LFT & DE-LFT \\
Metric & RMSE & MAE & RMSE & MAE & RMSE & MAE \\
\midrule
Yelp & 1.0186 & 0.7951 & 1.0012 & 0.7812 & 0.9724 & 0.7668 \\
CDs &0.8905 & 0.5971 & 0.8509  & 0.5853 & 0.8347 & 0.5734 \\
Network7a & 2.3138 & 1.5627 & 2.2965 & 1.5339 & 2.2787 & 1.5125 \\
Network6a & 1.3021 & 0.6569 & 1.2764 & 0.6417 & 1.2518 & 0.6183 \\
\bottomrule
\end{tabular}
\end{table}

\section{Conclusion}
This paper proposes DE-LFT, a Differential Evolution based hyperparameter learning framework for Latent Factorization of Tensors. By automatically optimizing the regularization coefficients of latent feature matrices during training, the proposed method improves the prediction accuracy of LFT models. Experimental results on four real-world datasets demonstrate that DE-LFT consistently achieves lower MAE and RMSE compared with manual tuning and grid search methods.

\bibliographystyle{IEEEtran}
\bibliography{refs}  

\end{document}